\documentclass[conference]{IEEEtran}
\IEEEoverridecommandlockouts
\usepackage{cite}
\usepackage{amsmath,amssymb,amsfonts}
\usepackage{algorithmic}
\usepackage{graphicx}
\usepackage{textcomp}
\usepackage{xcolor}

\usepackage{makecell}
\usepackage{url}
\usepackage{times}
\usepackage{epsfig}
\usepackage{amssymb}
\usepackage{multirow}
\usepackage[ruled,linesnumbered]{algorithm2e}
\usepackage{xcolor}

\def\BibTeX{{\rm B\kern-.05em{\sc i\kern-.025em b}\kern-.08em
    T\kern-.1667em\lower.7ex\hbox{E}\kern-.125emX}}
\begin{document}

\title{DBDH: A Dual-Branch Dual-Head Neural Network for Invisible Embedded Regions Localization}

\author{
\IEEEauthorblockN{Chengxin Zhao\textsuperscript{1}, Hefei Ling\textsuperscript{1,*}, Sijing Xie\textsuperscript{1}, Nan Sun\textsuperscript{1}, Zongyi Li\textsuperscript{1}, Yuxuan Shi\textsuperscript{1}, Jiazhong Chen\textsuperscript{1}}
\IEEEauthorblockA{\textsuperscript{1}\textit{School of Computer Science and Technology},
\textit{Huazhong University of Science and Technology},
Hubei, China \\
\{chengxinzhao, lhefei, xiesijing, sunnan, zongyili, shiyx, jzchen\}@hust.edu.cn
\thanks{*Corresponding author}}
}

\maketitle

\begin{abstract}
Embedding invisible hyperlinks or hidden codes in images to replace QR codes has become a hot topic recently. This technology requires first localizing the embedded region in the captured photos before decoding. Existing methods that train models to find the invisible embedded region struggle to obtain accurate localization results, leading to degraded decoding accuracy. This limitation is primarily because the CNN network is sensitive to low-frequency signals, while the embedded signal is typically in the high-frequency form.
Based on this, this paper proposes a Dual-Branch Dual-Head (DBDH) neural network tailored for the precise localization of invisible embedded regions. Specifically, DBDH uses a low-level texture branch containing 62 high-pass filters to capture the high-frequency signals induced by embedding. A high-level context branch is used to extract discriminative features between the embedded and normal regions. DBDH employs a detection head to directly detect the four vertices of the embedding region. In addition, we introduce an extra segmentation head to segment the mask of the embedding region during training. The segmentation head provides pixel-level supervision for model learning, facilitating better learning of the embedded signals.
Based on two state-of-the-art invisible offline-to-online messaging methods, we construct two datasets and augmentation strategies for training and testing localization models. Extensive experiments demonstrate the superior performance of the proposed DBDH over existing methods.
\end{abstract}


\section{Introduction}
\label{sec:intro}

Embedding QR codes allows us to transfer information from offline physical materials to digital devices online. The appearance of the QR code can disrupt the visual aesthetics of the host image, limiting its application in various scenarios. Recent works \cite{conf/cvpr/WengrowskiD19, conf/cvpr/TancikMN20, conf/mm/FangJMCZ22, conf/cvpr/JiaGZMZY22, journals/tcyb/JiaGCHMZY22} have proposed the invisible hyperlink/hidden code technology based on information hiding, with the goal of achieving seamless offline-to-online messaging while maintaining a positive user experience. These methods embed messages in images that are imperceptible to the human eye, while allowing us to recover the message from photos taken with a smartphone. 

\begin{figure*}[t]
	\begin{center}
		\includegraphics[width=0.9\linewidth]{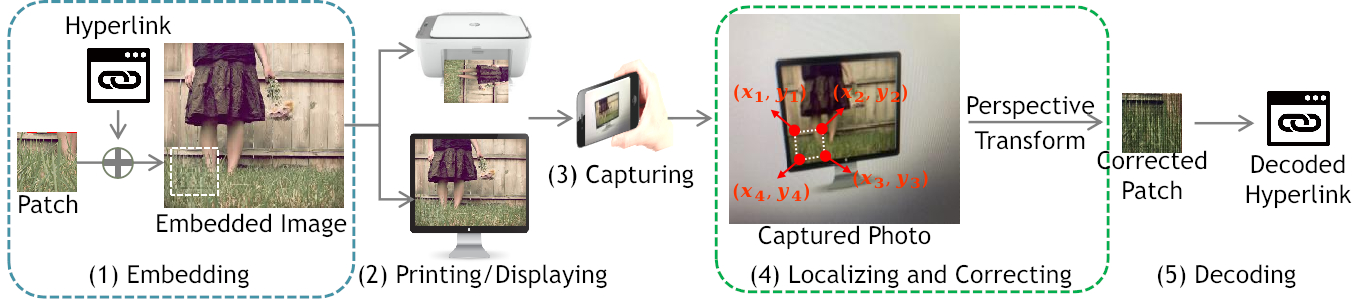}
	\end{center}
	\caption{The application pipeline of offline-to-online messaging in the print/screen-shooting scenario. The white dotted box indicates the embedded region.}
	\label{fig:pipeline}
\end{figure*}

Locating the embedded region is a prerequisite for decoding messages from the captured photo. As shown in Fig.\ref{fig:pipeline}, in the print/screen-shooting scenario, the message transmission process typically involves five steps: (1) embedding the message (e.g., a hyperlink) in the image; (2) printing or displaying on the screen; (3) capturing with a smartphone; (4) localizing the embedded region in the photo and correcting the geometric distortion caused by shooting; (5) decoding the message from the corrected image. In the above process, the geometric distortion introduced by the shooting process perturbs the relative position of the image, making the localization of the invisible embedded region a primary task before decoding. Existing methods \cite{conf/cvpr/WengrowskiD19, conf/cvpr/TancikMN20, conf/mm/FangJMCZ22, conf/cvpr/JiaGZMZY22, journals/tcyb/JiaGCHMZY22} adopt preprocessing to localize the four vertices of the embedded region, so that they can correct the geometric distortion by perspective transformation.

The localization performance is critical to the success of decoding. Early works \cite{journals/tifs/FangZZCY19, conf/cvpr/WengrowskiD19, journals/tcsv/FangZMZSCY20} obtain the vertices of the embedded region by the human eye, which has proven to be impractical in real-world applications. Later works \cite{journals/tmm/FangCWMLZZY22, journals/tcyb/JiaGCHMZY22} draw visible markers around the vertices of the embedded region to achieve automatic detection. However, this strategy introduces additional visual artifacts that break the imperceptibility. Recently, non-referencing localization methods based on self-supervised learning have been proposed. StegaStamp \cite{conf/cvpr/TancikMN20} first uses the BiSeNet \cite{conf/eccv/YuWPGYS18} to segment the embedded region, then the vertices are obtained by fitting a quadrilateral convex hull to the segmentation mask. Invisible Markers \cite{conf/cvpr/JiaGZMZY22} simplify the localization process by using an HRNet \cite{journals/pami/00010CJDZ0MTW0X21} to detect the vertices directly. Existing methods automate the localization process, however, they do not account for the differences between the embedded signal and the general visual signal. CNN-based models are typically sensitive to low-frequency signals \cite{conf/cvpr/0007QSWCR20}, while the invisible embedded signal is usually in high-frequency form \cite{conf/nips/ZhangBKSK20, conf/iccv/Jing0XWG21}. This causes these methods to perform poorly in localization. 

To address this problem, this paper proposes a novel Dual-Branch Dual-Head (DBDH) neural network tailored for accurate localization of invisible embedded regions. Inspired by steganalysis work \cite{journals/tifs/FridrichK12, conf/ih/SongLYLZ15}, DBDH includes a low-level texture branch that uses 62 carefully designed high-pass filters to capture high-frequency components in the image. It also uses a high-level context branch to extract discriminative features between the embedded and normal regions in a large visual receptive field. At the top of DBDH, there are two heads to localize the embedded region: a vertex detection head to directly detect the four vertices, and a segmentation head to predict the mask of the embedded region. The segmentation head is used during training only, which provides region-wise supervision to learn the entire embedding signal. 

The main contributions of this paper are as follows:

\begin{itemize}

\item We propose a novel Dual-Branch Dual-Head network for accurate localization of invisible embedded regions. It incorporates a low-level branch that uses sophisticated high-pass filters to capture the high-frequency signals caused by embedding. An auxiliary segmentation head is employed to provide region-wise supervision for better learning of the embedded feature.

\item We construct two embedded image datasets based on two state-of-the-art invisible offline-to-online messaging schemes. To improve model robustness, we introduce two image augmentation strategies during training. They simulate the print-shooting and screen-shooting processes, respectively.

\item We demonstrate the superior localization performance and the robustness of the proposed DBDH through extensive experiments.
\end{itemize}

\begin{figure*}[t]
	\begin{center}
		\includegraphics[width=0.95\linewidth,height=0.5\linewidth]{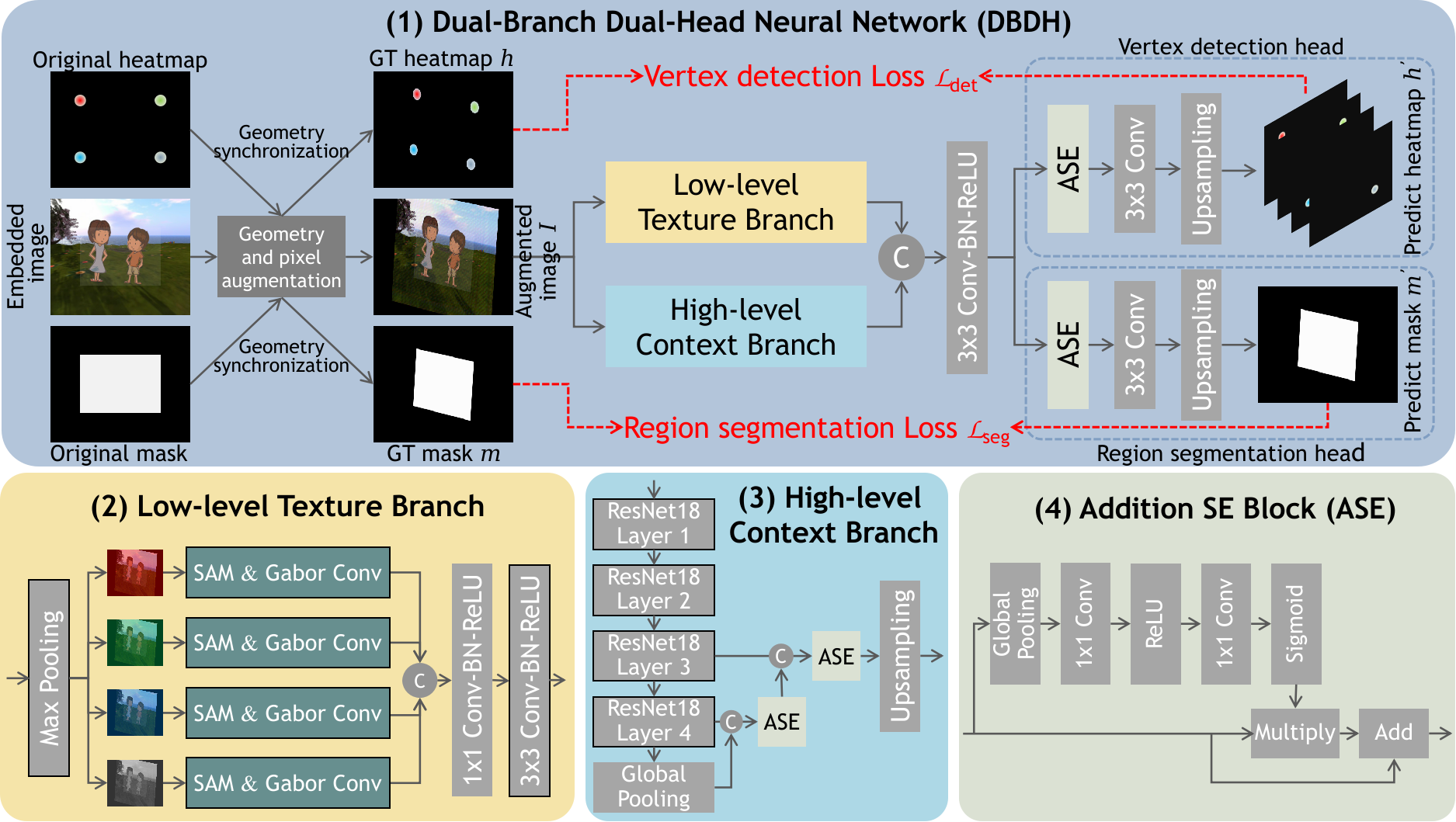}
	\end{center}
	\caption{Overall architecture of the proposed Dual-Branch Dual-Head (DBDH) network.  Dual branches: the low-level texture branch uses fixed SRM and Gabor kernels to obtain high-frequency components of the augmented image, while the high-level context branch uses a ResNet18 to obtain the discriminative feature between the embedded and the normal region. Dual head: the vertex detection head detect the four vertices of the embedded region, and the region segmentation head outputs the embedded region's mask, which serves as an auxiliary supervision. Blocks with a black border indicate that their stride is 2.}
	\label{fig:network}
\end{figure*}

\section{Related Work}

\subsection{Robust watermarking for offline-to-online messaging}
%

Based on camera-shooting-resilient watermarking, it is possible to embed invisible messages in offline photos to transmit them online. Traditional cross-media watermarking methods \cite{journals/tifs/FangZZCY19, journals/tcsv/FangZMZSCY20,  journals/tmm/FangCWMLZZY22} focus on improving the robustness against pixel-level distortions. They require manual localization of the embedded region from the captured photos. Recently, deep watermarking methods based on the Encoder-Noiselayer-Decoder (END) framework have become mainstream. By introducing the print/screen camera imaging process into the noise layer and optimizing the watermarking model end-to-end, many advanced offline-to-online messaging schemes \cite{conf/cvpr/WengrowskiD19, conf/cvpr/TancikMN20, journals/tcyb/JiaGCHMZY22, conf/cvpr/JiaGZMZY22} have emerged. 

Due to the imperceptibility of the watermark signal, the localization of the embedded region remains a challenge. StegaStamp  \cite{conf/cvpr/TancikMN20} first adopts the learning-based localization scheme, and trains a BiSeNet \cite{conf/eccv/YuWPGYS18} to segment the embedded region. Invisible Markers suggests using the HRNet \cite{journals/pami/00010CJDZ0MTW0X21} to directly detect the four vertices of the embedded region.  However, CNN-based models excel at extracting low-frequency signals \cite{conf/cvpr/0007QSWCR20}, they struggle with the invisible high-frequency signals introduced by the embedding \cite{conf/nips/ZhangBKSK20, conf/iccv/Jing0XWG21}, resulting in poor localization performance. Therefore, this paper utilizes carefully designed high-pass filters in addition to trainable layers to explicitly capture high-frequency signals. This strategy allows for more precise modulation of the embedded signal, resulting in more accurate localization results.



\section{Method}
A novel Dual-Branch Dual-Head neural network designed for localization of invisible embedded regions is presented here, which is called DBDH. As shown in Fig.\ref{fig:network} (1), DBDH contains two branches that are used to extract different levels of features. Two heads are placed at the top to output the location information of the embedded region. Details of the four main components are described below.


\subsection{Low-level texture branch}
Given the image for localization, the low-level branch is designed to preserve the resolution of the embedded image to encode enough texture features.
Since the invisible embedded signal is typically in a high-frequency form \cite{conf/nips/ZhangBKSK20, conf/iccv/Jing0XWG21}, we design an \textit{SRM\&Gabor Conv} layer consisting of 62 high-pass filters to explicitly extract the shallow features of the embedded image. Based on previous steganalysis work, the adopted filters include 30 basic SRM kernels \cite{journals/tifs/FridrichK12} and 32 Gabor kernels \cite{conf/ih/SongLYLZ15}. These filters are carefully designed to extract rich anomalous signals in the relatively high frequency sub-bands. For full extraction, the R, G, B, and Y color channels of the input image are convolved using the \textit{SRM\&Gabor Conv} layer. The resulting feature maps are concatenated along the channel dimension. We then use a 1 $\times$ 1 convolutional block (i.e., \textit{Conv-BN-ReLU}) to compress the channel dimension of the fused feature map, followed by a 3 $\times$ 3 convolutional block with a stride of 2 to reduce the spatial scale. Fig.\ref{fig:network} (2) shows the structure of this texture branch.

\subsection{High-level context branch}
Within a large receptive field \cite{conf/cvpr/PengZYLS17}, the high-level context branch aims to capture discriminative features between the embedded and normal regions. To keep our model lightweight, we use ResNet18 \cite{conf/cvpr/HeZRS16} as the basic structure of this branch. Similar to BiSeNet \cite{conf/eccv/YuWPGYS18}, we add a global average pooling at the top of this branch to provide a full receptive field for model learning. In addition, we adopt the channel attention mechanism \cite{journals/pami/HuSASW20}, i.e., the Addition SE Block (ASE), to perform feature refinement in the last two layers. This helps the context branch learn more refined features to distinguish the embedded region from the normal region. Finally, we use a bilinear upsampling layer to interpolate the output feature map to be the same size as the texture branch outputs. Fig.\ref{fig:network} (3) shows the structure of the context branch and Fig.\ref{fig:network} (4) shows the adopted Addition SE Block (ASE) in detail.

\subsection{Vertex detection head}
The vertex detection head aims to detect the four vertices of the embedded region so that the geometric distortion introduced by shooting can be directly corrected by perspective transformation. Here, we adopt the keypoint detection scheme to output the vertex. Specifically, this detection head outputs a 4-channel heatmap, and each heatmap corresponds to a vertex. A heatmap is a two-dimensional array of the same size as the input image, where the value of each pixel indicates the confidence that the pixel belongs to the vertex. At inference time, we select the position with the highest confidence as the predicted vertex coordinate. 

The features from the low-level texture branch and high-level context branch are merged using a 3 $\times$ 3 \textit{Conv-BN-ReLU} block. An ASE module is used in the detection head to further refine the merged feature and select suitable features for the detection task. The detection result is outputted using a 3 $\times$ 3 convolutional layer, then upsampled to match the input size.

\subsection{Region segmentation head}
The segmentation head is responsible for outputting the mask of the embedded region. It serves as a complementary branch to provide finer-grained supervision for learning embedding features. With the detection head, it is challenging for the network to learn robust features indicating embedded regions through vertex supervision only. Therefore, we incorporate this segmentation task to guide the model to modulate the entire embedding signal. 

The segmentation head has the same structure as the localization head, except that it outputs a single-channel segmentation mask. It is worth noting that the segmentation head is only used during training and does not add any computational overhead during inference.

\begin{figure*}[t]
	\begin{center}
		\includegraphics[width=0.99\linewidth]{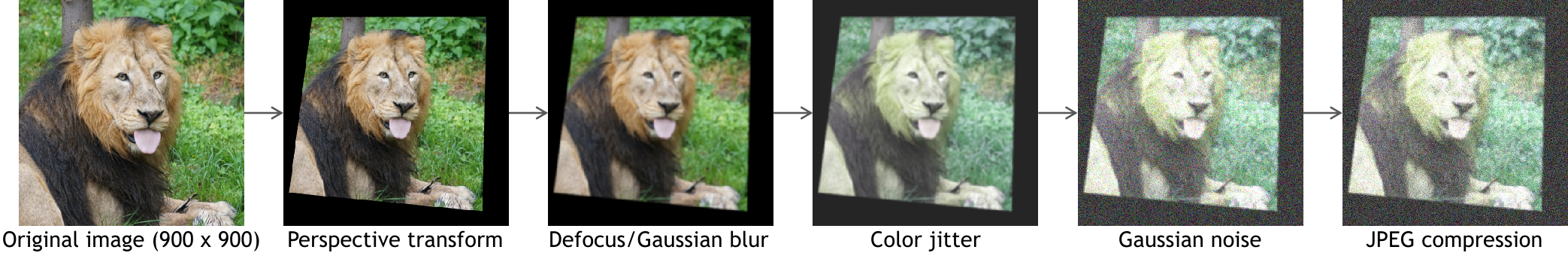}
	\end{center}
	\caption{Augmentation strategy for simulating the print-shooting process, which is called Aug-SS.}
	\label{fig:aug-ss}
\end{figure*}

\begin{figure}[t]
	\begin{center}
		\includegraphics[width=0.99\linewidth]{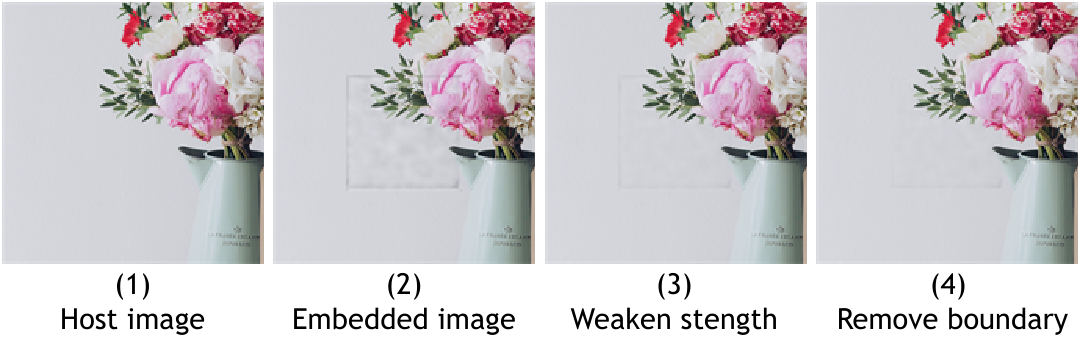}
	\end{center}
	\caption{Post-processes of the WM-SS dataset. It improves the average PSNR of WM-SS to 40.06 dB.}
	\label{fig:wm-ss}
\end{figure}

\subsection{Training Strategy}
To improve localization robustness in practice, we apply both geometry and pixel-level distortions to augment the embedded image during training. Since the offline-to-online messaging models have different robustness requirements, we introduce the augmentation strategies in the Datasets and augmentations section \ref{sec:Datasets}. Here, we directly report the training loss for the detection and segmentation tasks, respectively.

\textbf{Vertex detection Loss.} 
The location information of each vertex is represented by a heatmap, where a 2D Gaussian distribution ($\sigma=5$) is centered at that position. For clarity, we plot the four vertices in a single heatmap, as shown in Fig.\ref{fig:network} (1), and use different colors to distinguish them. For the augmented image $I \in \mathbb{R}^{C \times H \times W}$, we obtain its ground truth (GT) heatmap $h \in \mathbb{R}^{4 \times H \times W} $ by performing the same geometric transformation on the original heatmap during the augmentation. Here, we use FocalHeatmapLoss \cite{journals/ijcv/LawD20} as the vertex detection loss $\mathcal{L}_{\text{det}}$. Given the detection head prediction $h' \in \mathbb{R}^{4 \times H \times W}$, it is calculated as:
\begin{equation}
	\begin{gathered}
		\mathcal{L}_{\text{det}}=- \sum_{c=1}^4 \sum_{i=1}^H \sum_{j=1}^W l_{cij}^{\text{det}}, \\
		l_{cij}^{\text{det}} = 
		\begin{cases}
			\left(1-h'_{cij}\right)^\alpha \log \left(h'_{cij}\right) & \text {if } h_{cij}=1 \\ \left(1-h_{cij}\right)^\beta (h'_{cij})^\alpha \log \left(1-h'_{cij}\right) & \text {otherwise}
		\end{cases}
	\end{gathered}
\end{equation}
where $h'_{cij}$ is the prediction at the position $(c, i, j)$ and $h_{cij}$ is the corresponding ground truth, $\alpha$ and $\beta$ are the hyper-parameters of the focal loss. We set $\alpha=2$ and $\beta=4$ in all our experiments, following \cite{journals/ijcv/LawD20}. 

\textbf{Region segmentation Loss.} The ground truth mask $m \in \mathbb{R}^{H \times W}$ is obtained by performing the same geometric transformation as the augmented image. We use the binary cross entropy as the segmentation loss $\mathcal{L}_{\text{seg}}$. Given the segmentation head output $m' \in \mathbb{R}^{H \times W}$, it is calculated as:
\begin{equation}
		\begin{gathered}
				\mathcal{L}_{\text{seg}}= \frac{1}{HW} \sum_{i=1}^H \sum_{j=1}^W l_{ij}^{\text{seg}}, \\
				l_{ij}^{\text{seg}} = - [m_{ij} \log \left(m'_{ij}\right) + \left(1-m_{ij}\right) \log\left(1-m'_{ij}\right)], 
		\end{gathered}
\end{equation}
where $m'_{ij}$ is the prediction at the position $(i, j)$, and $m_{ij}$ is the corresponding ground truth.
The total loss of DBDH is the combination of the two loss terms, i.e.,
\begin{equation}
	\mathcal{L} = \lambda_{\text{det}} \mathcal{L}_{\text{det}} + \lambda_{\text{seg}}\mathcal{L}_{\text{seg}}
\end{equation}
where $\lambda_{\text{det}}$ and $\lambda_{\text{seg}}$ are the weight factors. We set $\lambda_{\text{det}}=1$ and $\lambda_{\text{seg}}=10$ to make the two are of the same order of magnitude.

\section{Experiments}
In this section, we first construct the training and test datasets based on two state-of-the-art offline-to-online messaging schemes. The detailed experimental settings are followed.  Then, we compare the proposed DBDH with existing methods. Finally, we demonstrate the effectiveness of our design through ablation studies.

\begin{figure*}[t]
	\begin{center}
		\includegraphics[width=0.95\linewidth]{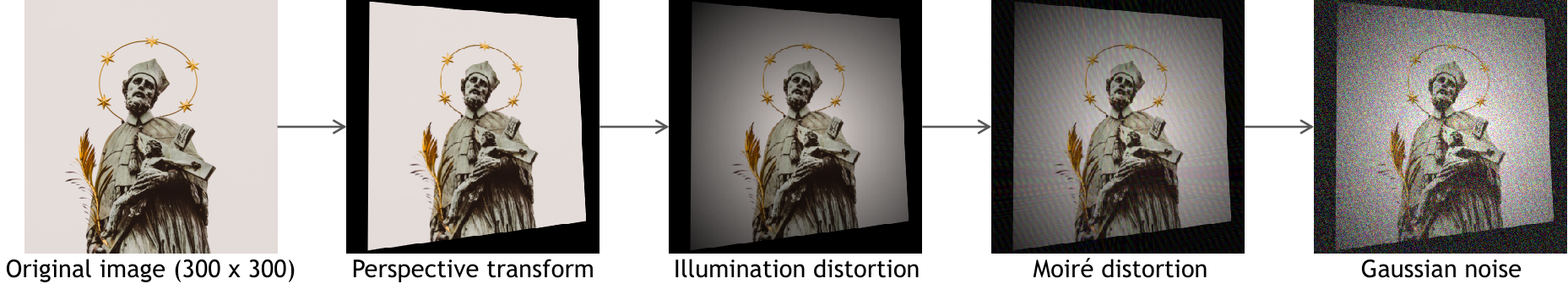}
	\end{center}
	\caption{Augmentation strategy for simulating the screen-shooting process, which is called Aug-PIMoG.}
	\label{fig:aug-pimog}
\end{figure*}

\subsection{Datasets and augmentations} \label{sec:Datasets}
The super-resolution image datasets DIV2K \cite{conf/cvprw/Agustsson_2017_CVPR_Workshops} and Flickr2K are used here to provide image data. They contain a total of 3550 2K-resolution images. We first resize these images to 1800 $\times$ 900, and then divide each image into three parts using a sliding window of size 900 $\times$ 900 with a stride of 450. The resulting 10,650 images are used as the host image for message embedding.

We adopt two state-of-the-art methods, i.e., StegaStamp \cite{conf/cvpr/TancikMN20} and PIMoG \cite{conf/mm/FangJMCZ22}, to embed messages into host images. The resulting embedded image datasets are denoted as WM-SS and WM-PIMoG, respectively. Since StegaStamp is designed for the print-shooting scenario and PIMoG is designed for the screen-shooting scenario, we develop two corresponding image augmentation strategies, namely, Aug-SS and Aug-PIMoG, to simulate the print-camera and screen-camera imaging processes, respectively. During training, the DBDH and other localization models are trained using augmented image data to improve their robustness.


\textbf{WM-SS.} WM-SS is the embedded image dataset generated by the pre-trained StegaStamp \cite{conf/cvpr/TancikMN20} model\footnote{https://github.com/tancik/StegaStamp}. The message is embedded in the central 400 $\times$ 400 patch of the host image. As shown in Fig.\ref{fig:wm-ss} (2), the initial embedded image has obvious visual artifacts, which is unaesthetic and easily leads to over-fitting of the localization model. Therefore, we scale down the pixel modification by a factor of 0.6 to weaken the embedding strength. Additionally, we replace the outermost 10 pixels of the embedded region with the original pixels, which removes the boundary artifacts. These post-processes increase the average PSNR of WM-SS from 34.2 dB to 40.06 dB. The processed results are shown in Fig.\ref{fig:wm-ss} (3)(4).


\textbf{Aug-SS.} Aug-SS is the augmentation strategy designed to simulate the distortion caused by the print-shooting process. Here, we adopt the noise layers used in StegaStamp to augment the embedded image in WM-SS. Specifically, we first warp the embedded image with a random perspective transform (scale $\le$ 0.3). This changes the relative position of the embedded region. Then we apply pixel-level distortions, including motion and defocus blur (kernel size $\in$ \{3, 5, 7\}), color jitter (brightness offset $\in [-0.3, 0.3]$, contrast offset $\in [0.5, 1.5]$, saturation offset $\in [0, 1]$, hue offset $\in [-0.2, 0.2]$), Gaussian noise ($\sigma \in [0, 0.2]$) and JPEG compression (quality factor $\in [50, 100]$). These are distortions that easily occur during the print-shooting process, which are shown in Fig.\ref{fig:aug-ss}.

\textbf{WM-PIMoG.} WM-PIMoG is the embedded image dataset generated by the pre-trained PIMoG \cite{conf/mm/FangJMCZ22} model\footnote{https://github.com/FangHanNUS/PIMoG}. Since PIMoG takes the image size of 128 $\times$ 128 as input, we center crop the original host image to 300 $\times$ 300 and take the central patch size of 128 $\times$ 128 as the embedded region. WM-PIMoG dataset achieves an average PSNR of 45 dB without any additional post-process.

\textbf{Aug-PIMoG.} Aug-PIMoG is the augmentation strategy designed to simulate the distortion caused by the screen-shooting process. We adopt the noise layers used in PIMoG to augment the embedded image in WM-PIMoG. As shown in Fig.\ref{fig:aug-pimog}, Aug-PIMoG involves four types of distortion: perspective transform (scale $\le$ 0.3), illumination distortion, moiré distortion, and Gaussian noise ($\sigma \in [0, 0.2]$). The illumination distortion multiplies the embedded image with a random illumination distribution map, and the moiré distortion is implemented by adding a random moiré pattern. Details of both can be found in PIMoG \cite{conf/mm/FangJMCZ22}. 

\subsection{Experimental settings}
For the datasets mentioned above, i.e., WM-SS and WM-PIMoG,  we randomly select 10,000 images as the training set, 300 images for validation and the remain 350 images consist the test set. During training, the batch size is set to 16 for WM-SS and 32 for WM-PIMoG. We train the proposed DBDH for 60 epochs by default. As for the gradient descent method, we use Adam \cite{journals/corr/KingmaB14} with a learning rate of 1e-3. The weight decay is set to 1e-5. 
During testing, we use the predicted vertices to localize the embedded region, and use the Intersection over Union (IoU, \%) between the localization result and the ground truth as the evaluation metric.

%

\begin{figure}[t]
	\begin{center}
		\includegraphics[width=0.99\linewidth]{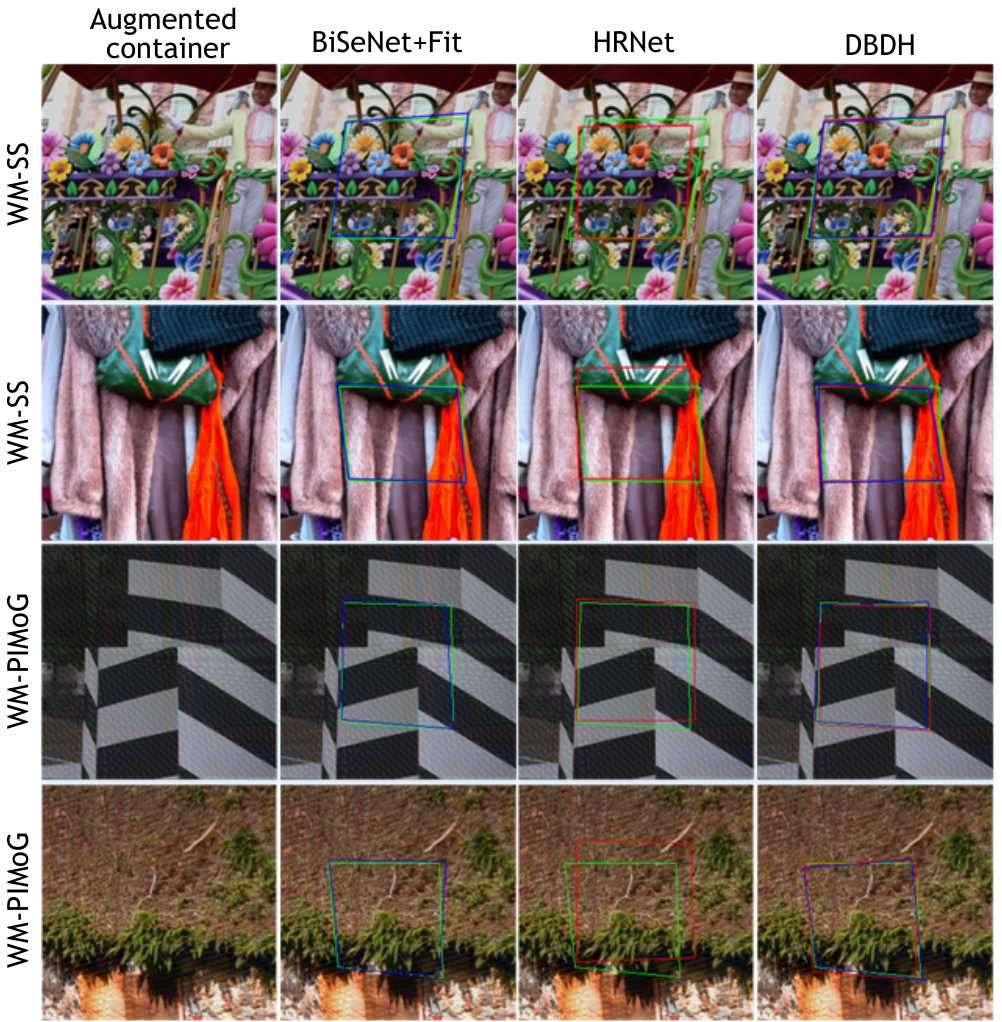}
	\end{center}
	\caption{Localization results under the combined distortion. The green rectangle represents the ground truth, the blue and red rectangles represent the localization results obtained by region segmentation and vertex detection, respectively.}
	\label{fig:comparison}
\end{figure} 

\begin{table*}[!h]
	\centering
	\caption{Comparison results. 'None' means no distortion during testing, 'combined' means the embedded image is distorted by all distortions. '\#Mult-Adds' refers to the number of multiply-add operations required by the model.}
	\resizebox{\linewidth}{!}{
		\begin{tabular}{c|c|cccccc|ccccc}
			\hline
			\multirow{2}{*}{Method} & \multirow{2}{*}{\#Mult-Adds} & \multicolumn{6}{c|}{IoU(\%) for WM-SS under Aug-SS} & \multicolumn{5}{c}{IoU(\%) for WM-PIMoG under Aug-PIMoG} \\ \cline{3-13} 
			& & None & Blur & Color jitter & Noise & JPEG & Combined & None & Illum. & Moire & Noise & Combined \\ \hline
			BiSeNet (StegaStamp \cite{conf/cvpr/TancikMN20}) & 35.21G & 95.8 & 94.0 & 94.3 & 95.0 & 95.5 & 90.3 & 90.3 & 90.2 & 91.0 & 89.7 & 87.0 \\
			BiSeNet+Fit (StegaStamp \cite{conf/cvpr/TancikMN20}) & 35.21G & 93.1 & 92.4 & 92.3 & 92.9 & 93.1 & 87.3 & 89.3 & 89.0 & 89.9 & 87.3 & 85.2 \\
			HRNet (IM \cite{conf/cvpr/JiaGZMZY22}) & 100.7G & 65.1 & 65.1 & 65.2 & 65.1 & 65.0 & 65.2 & 74.0 & 74.1 & 74.1 & 74.0 & 74.1 \\
			DBDH (Ours) & 30.71G & \textbf{97.0} & \textbf{96.2} & \textbf{95.8} & \textbf{95.5} & \textbf{96.8} & \textbf{91.2} & \textbf{91.3} & \textbf{91.0} & \textbf{92.1} & \textbf{89.8} & \textbf{87.6}\\ \hline
		\end{tabular}
	}
	\label{tab:comparison}
\end{table*}

\begin{table*}[!ht]
	\centering
	\caption{Ablation results. '$\times$' indicates that the texture branch is replaced with three stacked convolutional blocks. \\ '-' indicates that the adopted \textit{SRM\&Gabor Conv} layers are randomly initialized and trainable.}
	\resizebox{\linewidth}{!}{
		\begin{tabular}{c|c|cc|cccccc|ccccc}
			\hline
			\multirow{2}{*}{ID} & \multirow{2}{*}{\makecell[c]{Texture\\Branch}} & \multirow{2}{*}{\makecell[c]{Detection\\Head}} & \multirow{2}{*}{\makecell[c]{Segmentation\\Head}} & \multicolumn{6}{c|}{IoU(\%) for WM-SS under Aug-SS} & \multicolumn{5}{c}{IoU(\%) for WM-PIMoG under Aug-PIMoG} \\ \cline{5-15}
			& & & & None & Blur & Color jitter & Noise & JPEG & Combined & None & Illum. & Moire & Noise & Combined \\ \hline
			1 & $\times$ & \checkmark &  & 96.7 & 95.7 & 95.1 & 95.3 & 96.4 & 89.9 & 90.0 & 89.3 & 91.4 & 89.0 & 87.0 \\
			2 & $\times$ & \checkmark & \checkmark & 97.4 & 96.4 & 96.5 & 95.6 & 97.3 & 90.5 & 90.6 & 90.0 & 91.3 & 89.3 & 87.3 \\
			3 & - & \checkmark & \checkmark & 97.2 & 95.8 & 96.0 & 95.6 & 97.1 & 89.9 & 89.3 & 89.0 & 90.7 & 89.0 & 86.4 \\
			4 & \checkmark & \checkmark & \checkmark & \textbf{97.6} & 96.2 & \textbf{96.8} & 95.5 & \textbf{96.8} & \textbf{91.2} & \textbf{91.3} & \textbf{91.0} & \textbf{92.1} & \textbf{89.8} & \textbf{87.6} \\ \hline
		\end{tabular}
	}
	\label{tab:ablation}
\end{table*}

\subsection{Comparison with other methods}
\textbf{Baseline.} We use the two localization models adopted in Stegatamp \cite{conf/cvpr/TancikMN20} and Invisible Markers \cite{conf/cvpr/JiaGZMZY22} as our baseline. StegaStamp trains a segmentation model based on BiSeNet \cite{conf/eccv/YuWPGYS18} to predict the mask of the embedded region. The four vertices are then obtained by fitting a quadrilateral convex hull to the segmentation result. We use BiSeNet to denote the direct localization performance based on the segmentation model and BiSeNet+Fit to denote the performance of the fitted quadrilateral. Invisible Markers uses a HRNet \cite{journals/pami/00010CJDZ0MTW0X21} to detect the four vertices of the embedded region directly, we denote this method as HRNet.

\textbf{Model complexity.} Message retrieval requires efficient localization of the embedded region. We first compare the proposed DBDH with other methods in terms of computational cost and inference time. As shown in Table \ref{tab:comparison}, DBDH requires about 30.71 billion multiply-add operations, which is the lowest among these methods. When the input image size is 900 $\times$ 900 and the model is inferred on an NVIDIA-1080Ti, both DBDH and BiSeNet take about 25 ms, significantly less than the 78 ms required by HRNet. These results demonstrate the localization efficiency of DBDH.

\textbf{Localization performance.} As shown in Table \ref{tab:comparison}, although BiSeNet shows comparable performance to DBDH based on the segmentation results, the quadrilateral convex hull fitting process introduces extra bias, resulting in a decrease of IoU. Specifically, for the combined distortions, the vertex fitting process causes a 3\% decrease in WM-SS dataset and a 1.8\% decrease in WM-PIMoG dataset. Compared to the detection-based HRNet, the proposed DBDH achieves significant progress in both computational cost and localization accuracy. We attribute this to the reason that the supervision based on four vertices is too weak for the HRNet to learn robust features for localization, thus leading to a collapse of the model training. The qualitative comparison results are shown in Fig.6, which illustrates the superiority of DBDH.

\subsection{Ablation study}

We conduct four ablation experiments to validate the effectiveness of the low-level texture branch and the segmentation head. As shown in Table \ref{tab:ablation}, ID-1 replaces the texture branch with three stacked convolutional blocks and uses the vertex detection head only. ID-2 is based on ID-1 and adopts the segmentation head. Based on ID-2,  ID-3 adopts the texture branch, but makes the \textit{SRM\&Gabor Conv} layer trainable. 

\textbf{The effectiveness of the texture branch.} Comparing the last three rows of Table \ref{tab:ablation} shows that replacing the texture branch with convolutional blocks or making the \textit{SRM\&Gabor Conv} layer trainable leads to a decreased performance for localization. This demonstrates the effectiveness of the proposed high-pass filter-based texture branch. 


\textbf{The effectiveness of the segmentation head.} The model can easily collapse without the segmentation head. To make ID-1 converge, we train the model with the segmentation head in the first epoch, then drop it and optimize the model with the detection head only. It can be seen that even ID-1 converges, its performance is behind ID-2 and ID-4. This confirms that incorporating an additional segmentation head can improve the localization performance of invisible embedded regions.

\section{Conclusion}
The paper proposes a Dual-Branch Dual-Head neural network specifically designed for the localization of invisible embedded regions. The low-level branch incorporates 62 high-pass filters to capture the texture features introduced by message embedding, while the high-level branch extracts global context features to discriminate the embedded region. DBDH adopts a dual-head structure: one vertex detection head directly outputs the vertex coordinates of the embedded region, and the other segmentation head predicts the embedded region mask during training. The additional region-wise supervision is validated as beneficial for stabilizing the training process and improving localization accuracy. Through extensive comparison and ablation experiments on the two constructed datasets, we validate the superiority of the proposed DBDH in both simulated print-shooting and screen-shooting scenarios.

\section*{Acknowledgment}
This work was supported in part by the Natural Science Foundation of China under Grant 61972169, in part by the National key research and development program of China(2019QY(Y)0202, 2022YFB2601802), in part by the Major Scientific and Technological Project of Hubei Province(2022BAA046, 2022BAA042), in part by the Research Programme on Applied Fundamentals and Frontier Technologies of Wuhan(2020010601012182) and the Knowledge Innovation Program of Wuhan-Basic Research, in part by China Postdoctoral Science Foundation 2022M711251.

\bibliographystyle{IEEEtran}
\bibliography{egbib}

\begin{thebibliography}{10}
\providecommand{\url}[1]{#1}
\csname url@samestyle\endcsname
\providecommand{\newblock}{\relax}
\providecommand{\bibinfo}[2]{#2}
\providecommand{\BIBentrySTDinterwordspacing}{\spaceskip=0pt\relax}
\providecommand{\BIBentryALTinterwordstretchfactor}{4}
\providecommand{\BIBentryALTinterwordspacing}{\spaceskip=\fontdimen2\font plus
\BIBentryALTinterwordstretchfactor\fontdimen3\font minus
  \fontdimen4\font\relax}
\providecommand{\BIBforeignlanguage}[2]{{%
\expandafter\ifx\csname l@#1\endcsname\relax
\typeout{** WARNING: IEEEtran.bst: No hyphenation pattern has been}%
\typeout{** loaded for the language `#1'. Using the pattern for}%
\typeout{** the default language instead.}%
\else
\language=\csname l@#1\endcsname
\fi
#2}}
\providecommand{\BIBdecl}{\relax}
\BIBdecl

\bibitem{conf/cvpr/WengrowskiD19}
E.~Wengrowski and K.~J. Dana, ``Light field messaging with deep photographic
  steganography,'' in \emph{{CVPR}}.\hskip 1em plus 0.5em minus 0.4em\relax
  Computer Vision Foundation / {IEEE}, 2019, pp. 1515--1524.

\bibitem{conf/cvpr/TancikMN20}
M.~Tancik, B.~Mildenhall, and R.~Ng, ``Stegastamp: Invisible hyperlinks in
  physical photographs,'' in \emph{{CVPR}}.\hskip 1em plus 0.5em minus
  0.4em\relax Computer Vision Foundation / {IEEE}, 2020, pp. 2114--2123.

\bibitem{conf/mm/FangJMCZ22}
H.~Fang, Z.~Jia, Z.~Ma, E.~Chang, and W.~Zhang, ``Pimog: An effective
  screen-shooting noise-layer simulation for deep-learning-based watermarking
  network,'' in \emph{{ACM} Multimedia}.\hskip 1em plus 0.5em minus 0.4em\relax
  {ACM}, 2022, pp. 2267--2275.

\bibitem{conf/cvpr/JiaGZMZY22}
J.~Jia, Z.~Gao, D.~Zhu, X.~Min, G.~Zhai, and X.~Yang, ``Learning invisible
  markers for hidden codes in offline-to-online photography,'' in
  \emph{{CVPR}}.\hskip 1em plus 0.5em minus 0.4em\relax {IEEE}, 2022, pp.
  2263--2272.

\bibitem{journals/tcyb/JiaGCHMZY22}
J.~Jia, Z.~Gao, K.~Chen, M.~Hu, X.~Min, G.~Zhai, and X.~Yang, ``{RIHOOP:}
  robust invisible hyperlinks in offline and online photographs,'' \emph{{IEEE}
  Trans. Cybern.}, vol.~52, no.~7, pp. 7094--7106, 2022.

\bibitem{journals/tifs/FangZZCY19}
H.~Fang, W.~Zhang, H.~Zhou, H.~Cui, and N.~Yu, ``Screen-shooting resilient
  watermarking,'' \emph{{IEEE} Trans. Inf. Forensics Secur.}, vol.~14, no.~6,
  pp. 1403--1418, 2019.

\bibitem{journals/tcsv/FangZMZSCY20}
H.~Fang, W.~Zhang, Z.~Ma, H.~Zhou, S.~Sun, H.~Cui, and N.~Yu, ``A camera
  shooting resilient watermarking scheme for underpainting documents,''
  \emph{{IEEE} Trans. Circuits Syst. Video Technol.}, vol.~30, no.~11, pp.
  4075--4089, 2020.

\bibitem{journals/tmm/FangCWMLZZY22}
H.~Fang, D.~Chen, F.~Wang, Z.~Ma, H.~Liu, W.~Zhou, W.~Zhang, and N.~Yu,
  ``{TERA:} screen-to-camera image code with transparency, efficiency,
  robustness and adaptability,'' \emph{{IEEE} Trans. Multim.}, vol.~24, pp.
  955--967, 2022.

\bibitem{conf/eccv/YuWPGYS18}
C.~Yu, J.~Wang, C.~Peng, C.~Gao, G.~Yu, and N.~Sang, ``Bisenet: Bilateral
  segmentation network for real-time semantic segmentation,'' in \emph{{ECCV}
  {(13)}}, ser. Lecture Notes in Computer Science, vol. 11217.\hskip 1em plus
  0.5em minus 0.4em\relax Springer, 2018, pp. 334--349.

\bibitem{journals/pami/00010CJDZ0MTW0X21}
J.~Wang, K.~Sun, T.~Cheng, B.~Jiang, C.~Deng, Y.~Zhao, D.~Liu, Y.~Mu, M.~Tan,
  X.~Wang, W.~Liu, and B.~Xiao, ``Deep high-resolution representation learning
  for visual recognition,'' \emph{{IEEE} Trans. Pattern Anal. Mach. Intell.},
  vol.~43, no.~10, pp. 3349--3364, 2021.

\bibitem{conf/cvpr/0007QSWCR20}
K.~Xu, M.~Qin, F.~Sun, Y.~Wang, Y.~Chen, and F.~Ren, ``Learning in the
  frequency domain,'' in \emph{{CVPR}}.\hskip 1em plus 0.5em minus 0.4em\relax
  Computer Vision Foundation / {IEEE}, 2020, pp. 1737--1746.

\bibitem{conf/nips/ZhangBKSK20}
C.~Zhang, P.~Benz, A.~Karjauv, G.~Sun, and I.~S. Kweon, ``{UDH:} universal deep
  hiding for steganography, watermarking, and light field messaging,'' in
  \emph{NeurIPS}, 2020.

\bibitem{conf/iccv/Jing0XWG21}
J.~Jing, X.~Deng, M.~Xu, J.~Wang, and Z.~Guan, ``Hinet: Deep image hiding by
  invertible network,'' in \emph{{ICCV}}.\hskip 1em plus 0.5em minus
  0.4em\relax {IEEE}, 2021, pp. 4713--4722.

\bibitem{journals/tifs/FridrichK12}
J.~J. Fridrich and J.~Kodovsk{\'{y}}, ``Rich models for steganalysis of digital
  images,'' \emph{{IEEE} Trans. Inf. Forensics Secur.}, vol.~7, no.~3, pp.
  868--882, 2012.

\bibitem{conf/ih/SongLYLZ15}
X.~Song, F.~Liu, C.~Yang, X.~Luo, and Y.~Zhang, ``Steganalysis of adaptive
  {JPEG} steganography using 2d gabor filters,'' in \emph{IH{\&}MMSec}.\hskip
  1em plus 0.5em minus 0.4em\relax {ACM}, 2015, pp. 15--23.

\bibitem{conf/cvpr/PengZYLS17}
C.~Peng, X.~Zhang, G.~Yu, G.~Luo, and J.~Sun, ``Large kernel matters - improve
  semantic segmentation by global convolutional network,'' in
  \emph{{CVPR}}.\hskip 1em plus 0.5em minus 0.4em\relax {IEEE} Computer
  Society, 2017, pp. 1743--1751.

\bibitem{conf/cvpr/HeZRS16}
K.~He, X.~Zhang, S.~Ren, and J.~Sun, ``Deep residual learning for image
  recognition,'' in \emph{{CVPR}}.\hskip 1em plus 0.5em minus 0.4em\relax
  {IEEE} Computer Society, 2016, pp. 770--778.

\bibitem{journals/pami/HuSASW20}
J.~Hu, L.~Shen, S.~Albanie, G.~Sun, and E.~Wu, ``Squeeze-and-excitation
  networks,'' \emph{{IEEE} Trans. Pattern Anal. Mach. Intell.}, vol.~42, no.~8,
  pp. 2011--2023, 2020.

\bibitem{journals/ijcv/LawD20}
H.~Law and J.~Deng, ``Cornernet: Detecting objects as paired keypoints,''
  \emph{Int. J. Comput. Vis.}, vol. 128, no.~3, pp. 642--656, 2020.

\bibitem{conf/cvprw/Agustsson_2017_CVPR_Workshops}
E.~Agustsson and R.~Timofte, ``Ntire 2017 challenge on single image
  super-resolution: Dataset and study,'' in \emph{The IEEE Conference on
  Computer Vision and Pattern Recognition (CVPR) Workshops}, July 2017.

\bibitem{journals/corr/KingmaB14}
D.~P. Kingma and J.~Ba, ``Adam: {A} method for stochastic optimization,'' in
  \emph{{ICLR} (Poster)}, 2015.

\end{thebibliography}
\end{document}